\documentclass{article} 

\usepackage{geometry}
\usepackage[square,sort,comma,numbers]{natbib}
\usepackage[parfill]{parskip}

\usepackage[utf8]{inputenc} 
\usepackage[T1]{fontenc}    
\usepackage{hyperref}       
\usepackage{url}            
\usepackage{booktabs}       
\usepackage{amsfonts}       
\usepackage{nicefrac}       
\usepackage{microtype}      
\usepackage{graphicx}
\usepackage{amsmath}
\usepackage{float}
\usepackage{booktabs}
\usepackage{tabularx}
\usepackage{multirow}
\usepackage{cleveref}
\usepackage{scrextend}
\usepackage[many]{tcolorbox}
\usepackage{enumitem}
\usepackage{xcolor}
\usepackage{algorithm}
\usepackage{algorithmic}
\usepackage{wrapfig}
\usepackage{bm}
\usepackage{subfigure}

\usepackage{bigstrut}
\usepackage{todonotes}

\makeatletter
\def\hlinewd#1{%
\noalign{\ifnum0=`}\fi\hrule \@height #1 %
\futurelet\reserved@a\@xhline}
\makeatother

\usepackage{xspace}

\newcommand{\name}{\textsc{LegalBench}\xspace}
\newcommand{\abercrombie}{{Abercrombie}\xspace}
\newcommand{\hearsay}{{Hearsay}\xspace}
\newcommand{\personaljdx}{{Personal Jurisdiction}\xspace}
\newcommand{\diversityjdx}{{Diversity Jurisdiction}\xspace}
\newcommand{\proa}{{PROA}\xspace}
\newcommand{\cuad}{{CUAD}\xspace}
\newcommand{\rulerecall}{{Rule QA}\xspace}
\newcommand{\ird}{{Intra-Rule Distinguishing}\xspace}

\newcommand{\curie}{GPT-3 (curie)~}
\newcommand{\davinci}{GPT-3 (davinci)~}
\newcommand{\jumbo}{J1-Jumbo~}
\newcommand{\grande}{J1-Grande~}
\newcommand{\jlarge}{J1-Large~}

\newcommand{\website}{\url{https://github.com/HazyResearch/legalbench}}
\newcommand{\ntasks}{44~}

\title{\name: Prototyping a Collaborative Benchmark for Legal Reasoning}

\usepackage{authblk}

\author[1, 2]{Neel Guha}
\author[1]{Daniel E. Ho}
\author[1]{Julian Nyarko}
\author[2]{Christopher Ré}

\affil[1]{Stanford Law School}
\affil[2]{Stanford Computer Science}

\begin{document}

\maketitle

\begin{abstract}
  Can foundation models be guided to execute tasks involving legal reasoning? We believe that building a benchmark to answer this question will require sustained collaborative efforts between the computer science and legal communities. To that end, this short paper serves three purposes. First, we describe how IRAC---a framework legal scholars use to distinguish different types of legal reasoning---can guide the construction of a Foundation Model oriented benchmark. Second, we present a seed set of \ntasks tasks built according to this framework. We discuss initial findings, and highlight directions for new tasks. Finally---inspired by the Open Science movement---we make a call for the legal and computer science communities to join our efforts by contributing new tasks. This work is ongoing, and our progress can be tracked here: \website.
\end{abstract}

\section{Introduction}\label{sec:introduction}
Advances in language modeling are changing how American lawyers and administrators envision the practice of law~\cite{frankenreiter2022natural}. In transactional settings, computational language tools are already being used in document review~\cite{epiq2021}, and have illustrated promise for more sophisticated tasks like due diligence~\cite{burrell2019}. In administrative and civil settings~\cite{glaze2021artificial,engstrom2020government}, many have identified the potential for computational tools to improve the accessibility of legal services~\cite{tito2017access, engstrom2020legal, thompson2015creating, wu2019ai}, thereby alleviating the United States' long standing access-to-justice crisis~\cite{legal2017justice}. Unsurprisingly, the high risk nature of these tools---and their position in society---has inspired calls for better, law specific evaluation and auditing regimes~\cite{engstrom2020algorithmic}.

The potential for impactful computational legal language tools has been magnified by the development of language Foundation Models (FM)---large scale models trained on massive corpora of text~\cite{bommasani2021opportunities}.\footnote{In referring to ``Foundation Models,'' we mean model families like GPT-3~\cite{brown2020language}, Jurassic~\cite{lieber2021jurassic}, PaLM~\cite{chowdhery2022palm}, and others. We are chiefly interested in the in-context learning capabilities of these models. Model families which require finetuning---like BERT and RoBERTA---lie beyond the scope of this work.} Even within the machine learning community, FMs have provoked discussions of a paradigm shift. Practitioners have shown how FMs can be guided to perform novel tasks without updating model parameters, using only task instructions and a few labeled demonstrations~\cite{brown2020language, radford2019language}. The effectiveness with which natural language can be used to instruct these models suggests a future in which non-technical experts in specialized domains may be capable of leveraging FMs~\cite{lee2022coauthor, wu2022promptchainer, jiang2022prompt}. 

Thus, a natural bridging question is: 
\begin{center}
\textbf{Can FMs be guided to execute tasks involving legal reasoning?}
\end{center}

Existing benchmarking efforts have failed to answer this question. Prior work on legal language evaluation focuses on masked-language models like BERT or RoBERTA, which require significant amounts of labeled task data to perform well, and cannot be guided via natural language~\cite{zheng2021does, chalkidis2021lexglue}. Prior work on benchmarking FMs has focused on datasets which lack the nuances of legal tasks (e.g. legal knowledge, complex multi-step reasoning, and specialized terminology)~\cite{srivastava2022beyond}. 

A benchmark for evaluating legal FM reasoning should help computer scientists and lawyers answer two questions:

\begin{enumerate}
    \item Which types of legal reasoning, if any, can the FM perform? Legal tasks are highly heterogeneous, and vary with regards to the knowledge and type of reasoning they require~\cite{ellsworth2005legal}. A benchmark should provide practitioners with a fine-grained vocabulary for describing the legal inferences that different FMs may be capable of.  
    \item What FM programming strategies work for law? Prior work has noted the importance of understanding how different prompting workflows can enable experts to understand and control the performance of FMs~\cite{wu2022promptchainer}. Thus, a benchmark should enable the study and development of law-specific prompting strategies, with special consideration to the properties that distinguish legal reasoning from other tasks.   
\end{enumerate}

Drawing inspiration from prior Open Science efforts~\cite{ching2018opportunities}, we believe that an interdisciplinary process is the best way to construct a benchmark capable of answering the questions above. Lawyers are best positioned to identify tasks capable of meaningfully measuring legal reasoning skills. Computer scientists are best positioned to determine how FMs can be applied and evaluated on these tasks. 

To that end, we present \name---the first steps towards constructing an open and collaborative legal reasoning benchmark. \name enables collaboration via a data-centric approach~\cite{datacentricai, liang2022advances}. Legal domain experts can construct and submit tasks (as small as 50-100 labelled samples) which evaluate a particular form of legal reasoning (e.g., determining where along the \textit{Abercrombie} distinctiveness spectrum a particular mark falls). These tasks will be available in a central repository, and used to evaluate new FMs. Our hope is that---over time---this growing collection of tasks can guide our understanding of the limitations and capabilities of FMs.

This paper makes the following contributions:
\begin{enumerate}
    \item  First, we describe how IRAC~\cite{stockmeyer_2021}—a framework legal scholars use to classify different types of legal reasoning—informs the construction of \name.
    \item Second, we present \ntasks initial tasks for \name, spanning 4 areas of law. We present preliminary findings on 5 different FMs which illustrate how these tasks may inform our understanding of FM legal reasoning.
    \item Finally, we make a call for additional tasks from the broader legal community, and highlight tasks which we believe are particularly important to capture.
\end{enumerate}

We hope that this benchmark will be interesting to a diverse set of communities. Legal practitioners may use these benchmarks to determine whether and where FMs may be integrated into existing workflows to improve outcomes for clients. Judges and other institutions responsible for regulating the practice of law can assess where performance is good enough so as to not compromise professional ethics. Computer scientists may benefit from studying the performance of these models in new domains, where distinct lexical properties and unique tasks may surface new insights.  

Before we progress further, we would like to note that the purpose of this work isn't to evaluate whether computational systems \textit{should} replace lawyers and legal officers, or to understand the positive and negative impacts of that replacement~\cite{engstrom2020legal}. Rather, our goal is to assess--from a technical perspective--the extent to which existing models can perform tasks that appear to require legal reasoning, which may be used to augment, educate, or assist attorneys, not replace them. Given the proliferation of computational legal tools, we believe that answering this question is vital for ensuring their safe and ethical usage.

\section{Quantifying legal reasoning through IRAC}
We use IRAC--a methodology for legal reasoning pervasive in the American legal system--to ground \name in two ways. First, we  distinguish between tasks that require legal reasoning in a classical sense, and those tasks that involve legal skills. This distinction allows us to align \name with the legal literature, which has noted that traditional legal reasoning constitutes only a subset of what lawyers do~\cite{ellsworth2005legal}. In short, \name encompasses both IRAC tasks--which capture this traditional legal reasoning--and ``classification'' tasks which correspond to more general purpose legally relevant classification tasks. Second, we use IRAC to further differentiate between different types of legal reasoning tasks.

\subsection{What is IRAC-based legal reasoning?}
We focus on legal reasoning in the context of American law, which typifies a common law system. In comparison to civil law systems (common in Europe), common law systems emphasize the role of prior cases (i.e. \textit{precedent}) in adjudicating disputes~\cite{lamond2006precedent}. Judges in common law systems are free to create rules via their decisions in cases. These decisions thus have significant legal relevance---judges in future cases are required to follow the reasoning used in precedent (a principle known as \textit{stare decisis}).

American legal scholars often describe ``legal reasoning'' as the process of determining the legal conditions that arise from a set of events or occurrences, with reference to both prior cases and codified laws~\cite{ellsworth2005legal}. A common framework for executing this type of legal reasoning is the \textbf{I}ssue, \textbf{R}ule, \textbf{A}pplication and \textbf{C}onclusion (\textbf{IRAC}) framework~\cite{iracwiki, stockmeyer_2021}. In this framework, legal reasoning decomposes into four sequential steps: 

\begin{enumerate}
    \item First, lawyers identify the legal issue in a given set of facts (\textbf{issue spotting}). An issue is often either (1) a specific unanswered legal question posed by the facts, or (2) an area of law implicated in the facts. Depending on the setting, a lawyer may be told the issue, or be required to \textit{infer} a possible issue. 
    \item Second, lawyers identify the relevant legal rules for this issue (\textbf{rule recall}). A rule is a statement of law which dictates the conditions that are necessary (or sufficient) for some legal outcome to be achieved. Rules can come from a variety of sources: the Constitution, federal and state statutes, regulations, and court opinions (case law). Importantly, rules can vary or differ between jurisdictions. Hence, the relevant rule in California might be different than the relevant rule in New York. 
    \item Third, lawyers apply these rules to the facts at hand (\textbf{application} or \textbf{analysis}). Application, or the analysis of rule applicability, consists of identifying those facts which are most relevant to the rule, and determining how those facts influence the outcome under the rule. Application can also involve referencing prior cases involving similar rules (i.e. \textit{precedent}), and using the similarities or differences to those cases to determine the outcome of the current dispute.
    \item Finally, lawyers reach a conclusion with regards to their application of law to facts, and determine what the legal outcome of those facts are (\textbf{conclusion}). 
\end{enumerate}

We illustrate this framework with a simple example. Suppose that BusinessMart---a large manufacturing corporation---is being sued by Amy in federal court on diversity jurisdiction.\footnote{Diversity jurisdiction gives federal courts the ability to hear cases between parties that are ``citizens'' of different states.} BusinessMart sells the majority of its goods in Texas, has its headquarters (where its CEO and board members sit and work) in California, and maintains a factory in Florida. A court is trying to determine---for the purposes of diversity jurisdiction---where BusinessMart's principal place of business is. 

\begin{itemize}
    \item \textbf{Issue}: Here, a narrow issue has been stated---where is BusinessMart's principal place of business?
    \item \textbf{Rule}: A lawyer would recognize that the most relevant rule here comes from the case \textit{Hertz Corp. v. Friend}\footnote{Hertz Corp. v. Friend, 559 U.S. 77 (2010).}, in which the Supreme Court determined ``that the phrase `principal place of business' refers to the place where the corporation's high level officers direct, control, and coordinate the corporation's activities.''
    \item \textbf{Application}: Applying this rule to the facts above yields two important observations. First, a corporation's CEO and board members are examples of high level officers referred to in \textit{Hertz} that control and conduct a company. Second, the place where BusinessMart's high level officers control the company is therefore California, as that is where the CEO and board sit and work.  
    \item \textbf{Conclusion}: Based on the chain of inference spelled out in the application stage, a lawyer would thus conclude that California is BusinessMart's principal place of business. 
\end{itemize}


\subsection{What is non-IRAC reasoning?}
The IRAC framework captures a formalistic notion of legal reasoning, centered around issues, rules, and fact patterns. But, many of the tasks that lawyers perform fall outside the bounds of IRAC. For instance:
\begin{itemize}
    \item When analyzing a contract, lawyers must identify the type and function of each clause.
    \item When writing a brief, lawyers must determine if the conclusion of a previous case entails a particular statement of argument.  
\end{itemize}

Though these tasks require the knowledge base and skillset of lawyers, they, arguably, do not always fit neatly within the IRAC framework. Hence, we consider these to be distinct from the examples offered in the previous section.

\subsection{\name, IRAC, and classification tasks}

We provide a conceptual overview of \name, illustrating how it categorizes different types of legal tasks.

\textbf{IRAC tasks}. IRAC offers a taxonomy of legal reasoning tasks. Namely, it allows us to define tasks as:
\begin{itemize}
    \item Issue tasks, which evaluate how well FMs can identify legal issues.
    \item Rule tasks, which evaluate how well FMs can recall legal rules.
    \item Application tasks, which evaluate how well FMs can apply specific legal rules to fact patterns.
    \item Conclusion tasks, which evaluate how well FMs can arrive at the correct legal outcome for a rule's application to a fact pattern.
\end{itemize}

We use IRAC in two ways. First, by associating different \name tasks to IRAC categories, we can draw higher level conclusions on the types of tasks that FMs can perform well. This allows us to balance general inferences about legal reasoning against rule-specific inferences. In initial experiments for example, we have already observed that FMs tend to perform significantly better at issue-spotting than application. 

Second, we use IRAC to guide evaluation of individual tasks. Following the IRAC definitions above, we use the following scheme when evaluating FM outputs: 
\begin{itemize}
    \item For Issue tasks, we evaluate whether the model output correctly states the issue, and does not output unrelated issues.
    \item For Rule tasks, we evaluate whether the model correctly states every element of the rule, and does not add additional language which is not present in the actual rule. 
    \item For Application tasks, we evaluate whether the model can generate an explanation of its answer that is legally coherent. We find that chain-of-thought prompts~\cite{wei2022chain} are a promising way to elicit such explanations. 
    \item For Conclusion tasks, we evaluate solely whether the model's determination of the outcome (i.e. typically a binary yes/no) is correct. The model is not required to generate an explanation.
\end{itemize}

\textbf{Non-IRAC tasks}. Additionally, \name also includes non-IRAC tasks, which we dub ``classification tasks.'' Though these tasks don't require IRAC-style reasoning, we believe it is important to include these tasks in \name for two reasons. First, these tasks are pervasive in the legal profession, and do require skills that lawyers are considered to distinctly hold. Second, practitioners have already highlighted these tasks as areas where computational tools could supplant or augment legal professionals. Therefore, understanding the performance of modeling approaches here is key.

\section{The initial \name tasks}
To illustrate the types of tasks which \name may contain---and to better guide those who may wish to contribute---we constructed \ntasks initial tasks for \name (Table \ref{tab:task_overview}). Our task construction is intentionally lightweight and intended to prioritize accessibility. All IRAC tasks follow a question-answer format, and are inspired by the types of questions that law students would answer in law school, or lawyers would answer in the course of their employment. The small size of these tasks means that any legal professional could design and construct a new task in a matter of hours.  

We provide a brief overview of the \ntasks tasks in this section. A more in-depth overview is available on our website.\footnote{\website} We provide an initial categorization of tasks to types (e.g. classification, issue-spotting, rule recall, application, and conclusion). However, we note that distinguishing between types can sometimes be difficult, and that differences of opinion may exist.

\begin{table}[h]
    \centering
    \begin{footnotesize}
    \begin{tabularx}{\textwidth}{XXXXXX}
        \textbf{Task/task family} & \textbf{Structure}  & \textbf{Language} & \textbf{Train size} & \textbf{Test size}    \\\toprule
        \abercrombie & 5-way classification  & Natural English & 5 & 95 \\ \midrule
        \cuad & 32 Binary classification tasks  & Contractual language & 5 per task & 95 per task \\ \midrule
        \diversityjdx & 6 Binary classification tasks  & Natural English & 5 per task & 100 per task\\ \midrule
        \hearsay & Binary classification  & Natural English & 5 & 95 \\ \midrule
        \personaljdx & Binary classification  &Natural English & 5 & 50 \\ \midrule
        \proa & Binary classification & Statutory text & 5 & 95 \\ \midrule
        \rulerecall & Generation &  Natural English & 0 & 50 \\\midrule
        \ird & 3-way classification & Natural English & 0 & 60 \\
        \bottomrule
    \end{tabularx}
    \caption{Task overview. Our \ntasks tasks group into 8 families, where certain families (e.g. CUAD) contain multiple tasks.}
    \label{tab:task_overview}
    \end{footnotesize}
\end{table}

\textbf{\cuad (Classification)}. We reformulate the CUAD~\cite{hendrycks2021cuad} dataset, which consists of annotated contracts from the EDGAR database, where annotations denote different types of clauses in each of the contracts. We select 32 of clause types, and create a binary classification task for each type, where the objective is to distinguish clauses of that type from other randomly sampled clauses.

\textbf{Rule QA (Rule)}. We construct a dataset of 50 questions, where each question asks the FM to produce information regarding a specific legal rule. We focus on questions which ask the FM to either (1) restate a rule, (2) identify where a rule is codified, or (3) list the factors employed in a particular rule.

\textbf{Abercrombie (Application and Conclusion)}. We test the Abercrombie distinctiveness rule, which specifies how ``distinct'' a product name (i.e. a ``mark'') is with respect to the product. The Abercrombie distinctiveness spectrum specifies five levels. The level assigned to a mark is used to determine whether that mark is eligible for protection. We create a dataset consisting of 95 product-mark pairs and their corresponding rating on the Abercrombie scale. The task is framed as a multiclass classification problem, where the FM is tasked with predicting the distinctiveness level for a candidate pair.

\textbf{Hearsay (Application and Conclusion)}. We test the hearsay rule, which governs when certain out-of-court statements may be inadmissible as evidence during a trial. Applying the hearsay rule requires understanding what a particular statement asserts, and how that assertion relates to an argument of fact. We manually generate 95 samples, where each sample describes a piece of evidence and a question of fact. The FM is tasked with classifying each sample according to whether it's hearsay.   

\textbf{Diversity Jurisdiction (Application and Conclusion)}. We test the diversity jurisdiction rule, which specifies that state claims may be brought in federal court provided those claims involve citizens of different states for certain minimum amounts. Applying this rule requires associating different parties to their states of citizenship and performing arithmetic operations over claim amounts. We manually define 6 templates, each testing the diversity jurisdiction rule under a different number of parties and claims. For each template, we programmatically generate 95 sentences describing a set of parties, their citizenships, and claim amounts. For each sample, the FM is tasked with determining whether diversity jurisdiction exists.  

\textbf{Personal Jurisdiction (Application and Conclusion)}. We test a simplified version of the personal jurisdiction rule, which governs when a court in a forum may exercise personal jurisdiction over a defendant for a particular legal claim. Applying this rule requires evaluating the extent of the defendant's interactions with the forum, and whether the claim is related to those contacts. We manually generate 50 fact patterns describing interactions between different individuals occurring in different states and a legal claim arising from those interactions. The task consists of determining whether---for a specified state---the claim against the defendant meets the contacts and nexus requirement to justify personal jurisdiction.

\textbf{PROA (Application and Conclusion)}. A private right of action (PROA) exists when a statute empowers an ordinary individual to legally enforce their rights by enabling them to file suit in court. We collect a sample of 95 statutes from different state codes, and manually annotate them as to whether they contain a private right of action. We evaluate whether a FM can apply the necessary rule to determine whether a statute creates a private right.

\textbf{\ird (Issue)}. We prototype an issue-spotting task by feeding the model with the fact patterns from the Hearsay, Personal Jurisdiction, and Abercrombie tasks. The model is tasked with mapping each fact pattern to its correct task name. This task is offered merely as an illustrative example of what an issue-spotting \textit{may} look like, and the types of reasoning it requires.

\section{Initial results and next directions}

\textbf{Initial results}. To illustrate the potential of \name, we conducted a preliminary investigation of 5 different FMs (varying in size and provider). We provide full results in the Appendix, and full prompts are available on the website. A sample of results for different models are provided in Table \ref{tab:few_shot_issue_conclusion}. We find that \name begins to provide us with an understanding of FM legal reasoning capabilities. For instance, we observe that: 

\begin{itemize}
    \item Across all task types, model size strongly correlates with performance, with larger models consistently outperforming smaller models.
    \item There are substantial performance differences for models with the same architectures. We attribute this to differences in pretraining regime and instruction finetuning. 
    \item FMs generally perform best at classification tasks, and worst at application tasks. Even when a model's conclusion performance on a task is high, its application performance on the same task is almost always significantly lower. 
    \item For some issue tasks, in-context demonstrations are not required, or only marginally improve performance. Task performance thus appears to be mostly driven by the task description used in the prompt.  
    \item Chain-of-thought prompting---which has been shown to help with algorithmic tasks like mathematical reasoning and code generation---substantially improves performance here as well. 
    \item FMs are generally better at performing tasks which can be reframed as information-extraction or relation-extraction tasks. 
\end{itemize}

\begin{table}[ht]
    \centering
    \begin{footnotesize}
    \begin{tabularx}{\textwidth}{lXXXXXX}
        Model & \cuad & \proa (C) & \diversityjdx (C) & \hearsay (C) & \personaljdx (C)& \abercrombie (C) \\\toprule
        Majority class & 0.34&	0.34&	0.36&	0.35&	0.37 &	0.07 \\ \midrule
        \davinci & 0.86	&0.86&	0.74&	0.68&	0.58&	0.42 \\ \midrule
        \curie&  0.50&	0.55&	0.47&	0.37&	0.43&	0.22  \\ \midrule
        \jumbo & 0.63&	0.35&	0.53&	0.32&	0.42&	0.21 \\ \midrule
        \grande & 0.44	&0.33&	0.55&	0.54 &	0.39&	0.13  \\ \midrule
        \jlarge & 0.59&	0.40&	0.35&	0.40&	0.37&	0.06 \\ \bottomrule
    \end{tabularx}
    \caption{FM few-shot performance on classification and conclusion (C) tasks. We report F1 (macro) for each classification and conclusion task. For \cuad and \diversityjdx,  we report the average F1 (macro) across the 32 and 6 tasks respectively.}
    \label{tab:few_shot_issue_conclusion}
    \end{footnotesize}
\end{table}

\textbf{Next steps}. As we've discussed above, \name is inspired by Open Science initiatives, and should be a community wide effort in which relevant stakeholders are encourage to participate in. We envision the following model of collaboration: 

\begin{enumerate}
    \item Any individual or group with relevant domain expertise may contribute one or more tasks to \name via our website. Tasks should ideally fall into the IRAC framework, and consist of at least 50 samples for inference.  
    \item When new FMs are released, we will run them on \name tasks and release results. 
    \item After \name has grown to encompass a sufficient number of tasks, we hope to---together with all task contributors to \name---collaboratively write a paper detailing the different tasks and discussing our findings~\cite{ching2018opportunities}. 
\end{enumerate}

For questions or more information, please see our website at \website.\\

\textbf{Acknowledgements}. We gratefully acknowledge the support of NIH under No. U54EB020405 (Mobilize), NSF under Nos. CCF1763315 (Beyond Sparsity), CCF1563078 (Volume to Velocity), and 1937301 (RTML); ARL under No. W911NF-21-2-0251 (Interactive Human-AI Teaming); ONR under No. N000141712266 (Unifying Weak Supervision); ONR N00014-20-1-2480: Understanding and Applying Non-Euclidean Geometry in Machine Learning; N000142012275 (NEPTUNE); NXP, Xilinx, LETI-CEA, Intel, IBM, Microsoft, NEC, Toshiba, TSMC, ARM, Hitachi, BASF, Accenture, Ericsson, Qualcomm, Analog Devices, Google Cloud, Salesforce, Total, the HAI-GCP Cloud Credits for Research program,  the Stanford Data Science Initiative (SDSI), and members of the Stanford DAWN project: Facebook, Google, and VMWare. The U.S. Government is authorized to reproduce and distribute reprints for Governmental purposes notwithstanding any copyright notation thereon. Any opinions, findings, and conclusions or recommendations expressed in this material are those of the authors and do not necessarily reflect the views, policies, or endorsements, either expressed or implied, of NIH, ONR, or the U.S. Government.

\bibliographystyle{plain}
\bibliography{sources}
\newpage
\appendix

\section{Additional results}
We evaluate 3 AI21 models~\cite{lieber2021jurassic}---\texttt{j1-jumbo} (178B), \texttt{j1-grande} (17B), and \texttt{j1-large} (7.5B)---and 2 OpenAI GPT-3 models~\cite{brown2020language}---\texttt{text-davinci-002} (176B parameters), and \texttt{text-curies-002} (6.7B parameters).\footnote{OpenAI has not published the exact parameter counts for these models. However, \cite{eleutherai}~infers them based on performance.} We evaluate in three regimes: zero-shot, few-shot, and chain-of-thought~\cite{wei2022chain}.\\

\begin{table}[ht]
    \centering
    \begin{footnotesize}
    \begin{tabularx}{\textwidth}{lXXXXXX}
        Model & \cuad  & \proa (C) & \diversityjdx (C) & \hearsay (C) & \personaljdx (C)& \abercrombie (C) \\\toprule
        Majority class & 0.34&	0.34&	0.36&	0.35&	0.37 &	0.07\\ \midrule
        \davinci & 0.77&	0.82&	0.62&	0.34&	0.53&	0.39 \\ 
        \jumbo & 0.38&	0.62&	0.40&	0.42&	0.52&	0.07	 \\ \bottomrule
    \end{tabularx}
    \caption{Zero-shot performance for classification and conclusion tasks. We report F1 (macro) for each task, except for \cuad and \diversityjdx, where we report the average F1 (macro) across the 32 and 6 tasks respectively.}
    \label{tab:zero_shot_all}
    \end{footnotesize}
\end{table}

\begin{table}[ht]
    \centering
    \begin{footnotesize}
    \begin{tabularx}{5cm}{lc}
        Model &  \rulerecall \\\toprule
        \jumbo & 0.46\\ \midrule
        \davinci & 0.48\\ \midrule
        \grande & 0.32 \\ \midrule
        \curie&  0.06 \\ \midrule
        \jlarge &	 0.24 \\ \bottomrule
    \end{tabularx}
    \caption{FM performance for \rulerecall. We report accuracy for each model.}
    \label{tab:rule_performance}
    \end{footnotesize}
\end{table}

\begin{table}[ht]
    \centering
    \begin{footnotesize}
    \begin{tabularx}{5cm}{lc}
        Model &  \ird \\\toprule
        \jumbo & 0.16 \\ \midrule
        \davinci &  0.43\\ \midrule
        \grande &  0.61\\ \midrule
        \curie&  0.16  \\ \midrule
        \jlarge & 0.23	  \\ \bottomrule
    \end{tabularx}
    \caption{FM performance for \ird. We report F1 (macro) for each model.}
    \label{tab:issue_performance}
    \end{footnotesize}
\end{table}

\begin{table}[ht]
    \centering
    \begin{footnotesize}
    \begin{tabularx}{\textwidth}{lXXXXX}
        Model &  \proa (C) & \diversityjdx (C) & \hearsay (C) & \personaljdx (C)& \abercrombie (C) \\\toprule
        \jumbo & 0.76 & 0.66 & 0.54 & 0.54& 0.34	 \\ \midrule
        \davinci & 0.92 & 0.78 & 0.73 & 0.79& 0.54\\ \bottomrule
    \end{tabularx}
    \caption{FM performance with chain-of-thought prompts for conclusion tasks. We report F1 (macro) for each task, and average across the 6 subtasks for \diversityjdx. }
    \label{tab:explanation_performance}
    \end{footnotesize}
\end{table}

\begin{table}[ht]
    \centering
    \begin{footnotesize}
    \begin{tabularx}{\textwidth}{lXXXXXX}
        Model & \proa (A) & \diversityjdx (A) & \hearsay (A) & \personaljdx (A)& \abercrombie (A) \\\toprule
        \jumbo & 0.41	& 0.66 & 0.32 & 0.2 & 0.39	 \\ \midrule
        \davinci & 0.83 & 0.78 & 0.62 & 0.4 & 0.48	 \\ \bottomrule
    \end{tabularx}
    \caption{FM few-shot performance on application tasks. We report accuracy for each task, except for \diversityjdx, where we report the average across 6 subtasks.}
    \label{tab:application_f1}
    \end{footnotesize}
\end{table}


\end{document}